\documentclass[letterpaper, 10 pt, conference]{ieeeconf}  

\usepackage{graphicx}
\usepackage{epstopdf}
\usepackage{etoolbox}
\usepackage{cite}
\usepackage{picinpar}
\usepackage{amsmath}
\usepackage{url}
\usepackage{flushend}
\usepackage{textcomp}
\usepackage{bm}

\IEEEoverridecommandlockouts                              

\title{\LARGE \bf
Development of Soft Tactile Sensor for Force Measurement \\ and Position Detection
}

\author{Wu-Te Yang$^{1}$, Zhian Kuang$^{2}$, Changhao Wang$^{1}$ and Masayoshi Tomizuka$^{1}$

\thanks{$^{1}$ The authors are with Department of Mechanical Engineering,
        University of California, Berkeley, 2103 Etcheverry Hall, CA 94720, USA
        {\tt\small wtyang, changhaowang, tomizuka@berkeley.edu}}%
\thanks{$^{2}$Zhian Kuang is with the Research Institute of Intelligent Control and Systems, Harbin Institute of Technology, Harbin 150001, P.R. China~{\tt\small zhiankuang@berkeley.edu}}%
}

\begin{document}

\maketitle
\thispagestyle{empty}
\pagestyle{empty}

\begin{abstract}
As more robots are implemented for contact-rich tasks, tactile sensors are in increasing demand. For many circumstances, the contact is required to be compliant, and soft sensors are in need. This paper introduces a novelly designed soft sensor that can simultaneously estimate the contact force and contact location. Inspired by humans' skin, which contains multi-layers of receptors, the designed tactile sensor has a dual-layer structure. The first layer is made of a conductive fabric that is responsible for sensing the contact force. The second layer is composed of four small conductive rubbers that can detect the contact location. Signals from the two layers are firstly processed by Wheatstone bridges and amplifier circuits so that the measurement noises are eliminated, and the sensitivity is improved. 
An Arduino chip is used for processing the signal and analyzing the data. The contact force can be obtained by a pre-trained model that maps from the voltage to force, and the contact location is estimated by the voltage signal from the conductive rubbers in the second layer. In addition, filtering methods are applied to eliminate the estimation noise. Finally, experiments are provided to show the accuracy and robustness of the sensor. Videos can be found in the link below\footnote[3]{\url{https://drive.google.com/drive/folders/1MRG3pWKSnkXq2HXEzWZQiGsa06J5JUbh?usp=sharing}}.

\end{abstract}

\section{INTRODUCTION}

With recent developments in soft robotics~\cite{c1,c2,c3}, robot arms equipped with soft grippers are able to achieve more human-like tasks such as grasping fragile objects like vegetables and fruits. Soft grippers are made of compliant materials, which have high elasticity and compliance. On the positive side, soft materials enable the robot to perform complex motions during operations. On the negative side, the flexibility makes the overall robot grasping and manipulation problems more challenging. To achieve robust performance in dexterous in-hand manipulation, soft grippers require soft touch sensors. The recent works have been discussed below.


\begin{figure}[http]
    \centering
    \includegraphics[width=220pt]{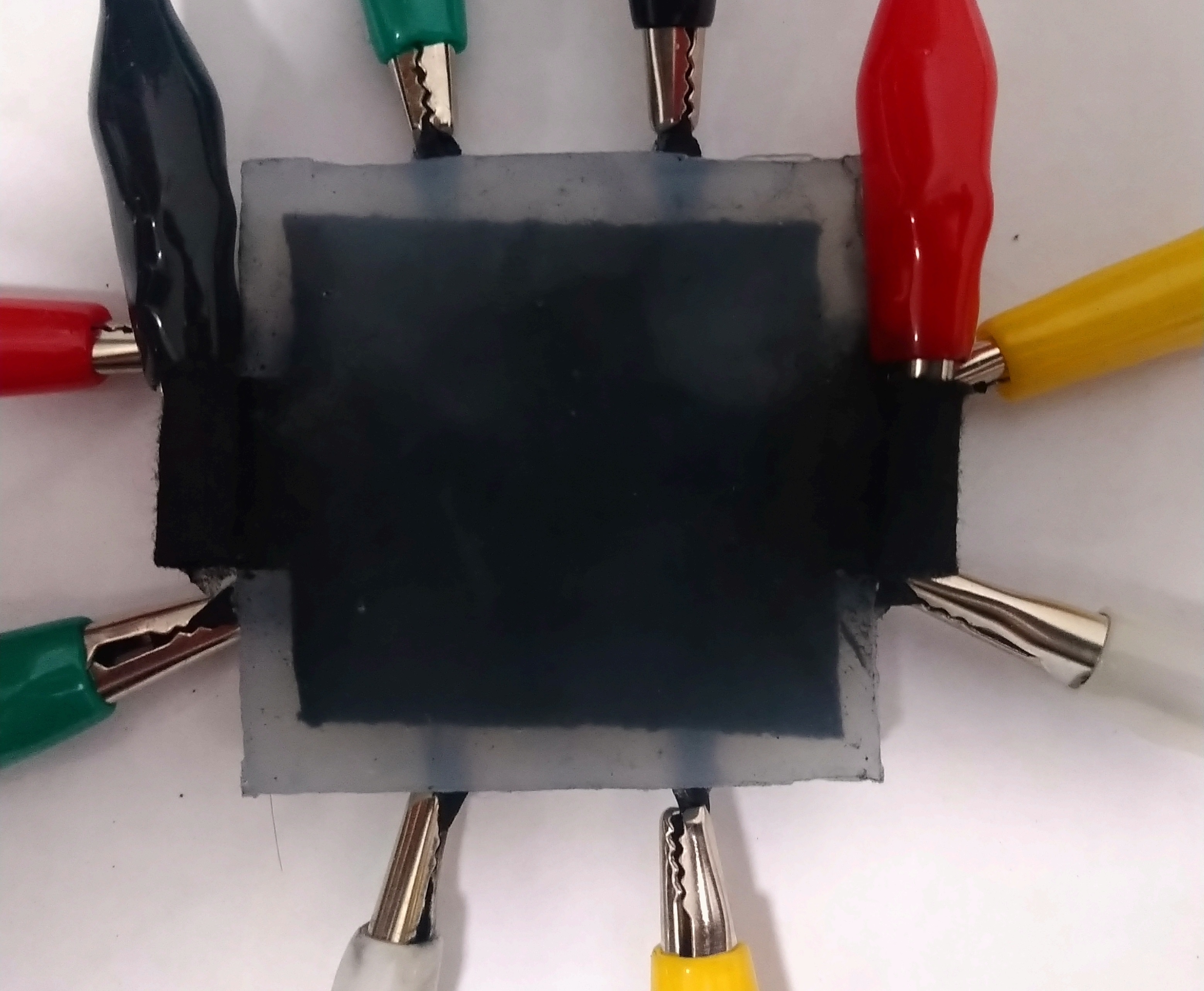}
    \caption{The appearance of the soft sensor.}
    \label{fig: Apparence}
\end{figure}


Capacitive sensors~\cite{c10,c11,c12,c13,c14,c15} are compliant and can be layout as sensor array to detect contact force in different positions~\cite{c12}. However, this type of sensors required complex circuits, which is hard to maintain in practice. Optical, ionic, and magnetic sensors have been developed recently. The vision-based sensors were embedded inside soft materials~\cite{c16,c17}. With machine learning algorithms, a model could be established to predict sensors' contact force or deformation. Nevertheless, the vision-based sensor might increase the costs, and the size of the vision sensors limits the size of soft sensors. In addition, Jamone et al.~\cite{c19} reported an idea about embedding magnets and Hall-effect sensors into soft material and formed soft sensors, which were integrated with the anthropomorphic robot hand. Unfortunately, the elasticity of soft sensors will be changed by embedded magnets and Hall-effect sensors inside.

Another type of sensors is resistive sensor. The traditional resistive sensors, such as the strain gauge, are high-cost and easy to fail, so the recent works used self-made resistive sensors. Singh et al.~\cite{c5} presented a design of a 3D printed soft resistive sensor. Hughes et al.~\cite{c6} introduced a soft sensor made by both sandwiched rubber and resistive line sensors, which was made of composite material. The line sensors were spanned in both x-direction and y-direction, and it could detect both deformation and contact positions. Ma et al.~\cite{c7} made a low-cost sensor with carbon powder. The resistive sensor was made of rubber and carbon powder. Shuichi et al.~\cite{c8} proposed a soft displacement method by painted conductive resin ink on a piece of rubber. Nassour et al.~\cite{c9} designed a soft sensor with two layers for the soft gripper. The first layer has multi-sensors that can detect force. The second one is the curvature sensor and was used to sense the curvature of the finger. However, those works might not be used to deal with manipulation tasks, which requires force estimation and contact location detection at the same time. 


\begin{figure}[http]
    \centering
    \includegraphics[width=230pt]{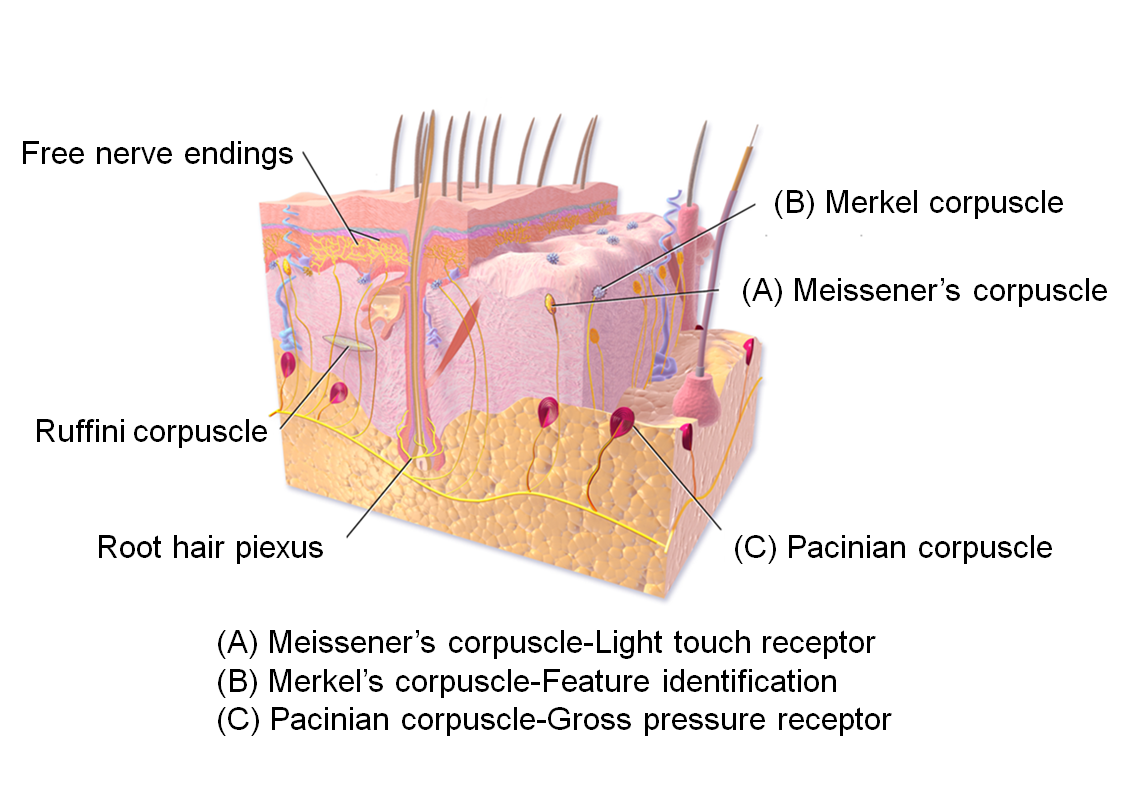}
    \caption{Humans' skin contain different kinds of receptors which are able to detect touch force and objects' features. Those receptors are Meissener's, Pacinian, and Merkel's corpuscle.~\cite{c18,c20}}
    \label{fig: human skin}
\end{figure}

The recently developed soft sensors usually could only perform a simple function, such as estimating contact force or deformation. The performance of them might not satisfy the requirements of robot grippers to do dexterous manipulation. By contrast, humans' hands are able to do complex grasping tasks without the help of vision because their skin has more than a single function such as force detection, slippery detection, and feature identification. Motivated from that, we designed a novel dual-layer soft sensor that is able to satisfy the need for dexterous manipulation. In~\cite{c6}, similar work presented that a multi-layer sensor, but it is unable to predict contact force. Another related work was proposed~\cite{c9}, but it cannot detect the contact location for further feature identification. 

Our soft tactile sensor contains two layers of sensors and can estimate contact force and contact position simultaneously. 
Resistive material is the ideal choice for both layers in this work since they are highly sensitive and small size, and the sensing circuit is relatively simple.  Due to the analysis of the sensing circuit, the output behavior is linear within a certain sensing range. The linear behavior of the sensor makes it easier to build the model and estimate the contact force accurately. In addition, since it is totally made of soft materials, it works consistently during deformation.

The remainder of this paper is organized as follows. Section II describes the design and fabrication of the soft sensor and the design of the sensing circuit. Section III models the sensor's behavior. Section IV demonstrates the experimental result, and Section V concludes the work.

\section{Design and Fabrication of Sensor}

Good touch sensitivity enables human to grasp objects that have complex geometric shapes even without vision.
Furthermore, when there is slippery, fingers can react within a few miniseconds~\cite{c20} and adjust the force to hold it stably. The main reason is that several types of receptors have different functions, as displayed in Fig.~\ref{fig: human skin}~\cite{c20}. Three of those receptors, including Meissener's, Pacinian, and Merkel's corpuscle, are responsible for light contact, gross contact, and feature identification, respectively. Those receptors enable our hands to perform complex manipulation tasks. The soft sensor should be similar to the skin of humans. Thus, we design the sensor that was inspired by the nature of humans, and will make robot grippers perform human-like tasks.

The sensor was designed to have dual-layers. The first one imitates the Meissener's and Pacinian corpuscles and is responsible for detecting contact force. The second one mimics the Merkel's corpuscle and takes the responsibility for locating the contact positions for further feature identification. For example, if the contact position is located and the contact pattern is identified as point contact, the object may be a sphere. If the contact pattern forms a line, the object might be a cylinder. The detailed design information will be discussed below.

\subsection{Design of the Force-measurement Layer}

The first layer (force-measurement layer) is similar to Meissner's and Pacinian corpuscles and will be applied to estimate the contact force. Like the human's skin, receptors are widely distributed over the finger. To achieve this, we selected conductive fabric for this layer, which could be cut into shapes that fit the possible contact area. The fabric is composed of 95 \% cotton and 5 \% spandex~\cite{c21}. After conductive ink was painted on its surface, the fabric became conductive and served as a soft resistive sensor. The key advantage of the conductive fabric is that it is highly elastic and can sustain large deformation during operations. It is also low-cost and easy to access. In the end, the sensor can be cut into any shape to fit into the sensor's size. In this work, the size of the fabric is 40 mm by 40 mm. It might be possible to be used to design artificial skin for detecting force in a large area in the future.

\begin{figure*}[h]
    \centering
    \includegraphics[width=1\textwidth]{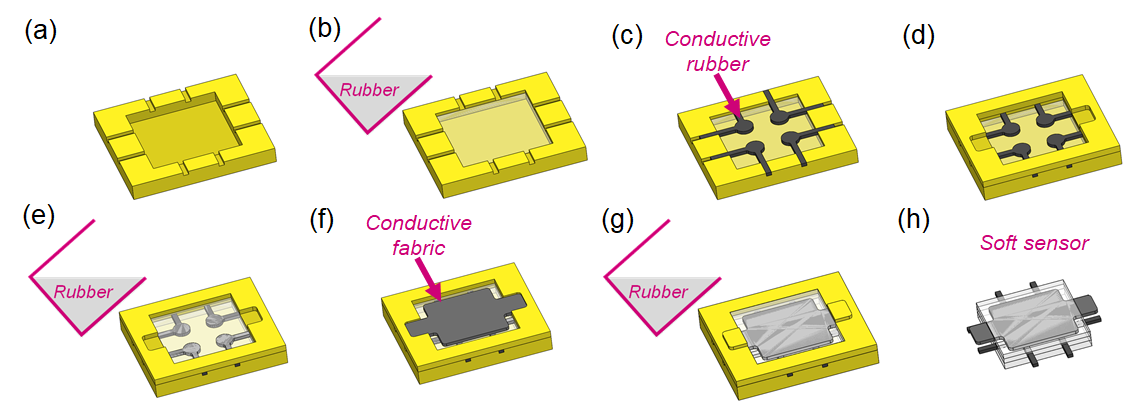}
    \caption{The fabrication process}
    \label{fig: process}
\end{figure*}

For the fabrication process, we soap a piece of fabric, which is made of cotton and spandex, in hot water at 60$^{o}$C for 30 minutes to relieve its residual stress. Then, the fabric will be hung for 3 hours to make it dry. By doing so, the elastic behaviors become more stable. 
Cotton and elastic fiber inside the fabric may behave unstable when undergoing cyclic stress because of the hysteresis effect~\cite{c22}. The main cause is the friction and residual stress inside the fabric. Soaping the fabric in hot water, which references the heat treatment for mechanical materials, can release its residual stress~\cite{c23}. The hysteresis effect problem can be reduced, and it is verified by real experiments. By contrast, the sensor behaves unstable when cyclic loads are applied without soaping in hot water. After this heat treatment-like process, the fabric was painted with conductive ink that is made of organic chemical and carbon powder with a ratio of 4:1 in weight.  The organic chemical is ethyl alcohol, whose concentration is 95$\%$. Then, the fabric is conductive and becomes a resistive sensor. 

When the fabric undergoes elastic deformation, the fabric's fibers will expand their length and reduce their cross-sectional areas. With the conductive ink on the surface, the fabric's resistance will change if elastic deformation occurs. Based on the definition of Ohm's law, the resistance R is defined as~\cite{c25}:
\begin{align}
R = \rho L/A,
\label{eqn: Ohm law}
\end{align}
where A is the cross-sectional area of the material, $\rho$ is the resistivity of the conductor, and L stands for the length. As the fibers of the fabric experience deformation caused by the contact force, the length L increases and cross-sectional A decreases because of the conservation of volume. The resistance of the fabric will increase based on (\ref{eqn: Ohm law}). By using this property, the conductive fabric can be utilized to measure the applied force.

\subsection{Design of the Position-detection Layer}

The second layer~(position-detection layer) is designed as a sensor array to detect the contact location. In order to achieve a high resolution, each sensor component should be small. The conductive fabric we used for the force-measurement layer is not a suitable material for this layer. When the fabric is cut into small pieces, each piece of fabric tends to make the structure loose and may break easily. Traditional resistive sensors, such as the strain gauge or the piezoelectric sensor, are made of metal. They are relatively weak and are easy to break~\cite{c4}. More importantly, when they are embedded inside soft materials, the sensor will become a composite material. Its mechanical properties will change (including Young's modulus) and makes the overall model analysis inaccurate. The metal has a large Young's modulus and will increase the sensor's Young's modulus~\cite{c24}. Thus, the sensor's elasticity and compliance will reduce and might not be able to sustain large deformation. Moreover, the decrease of elasticity and compliance will restrict the operations of the sensor. For example, if the grippers equipped with a low-compliance sensor undergo large deformation, the sensor cannot work consistently. 

One intuitive idea is to use soft materials with conductive ink to handle this problem. The soft sensor is made of silicone, Ecoflex{\textregistered} 00-30, and we use the same material for the sensor component in the position-detection layer. A set of molds is designed to make the sensor element. Liquid rubber, Ecoflex{\textregistered} 00-30, was poured into the mold. The liquid rubber was cured in 2 hours and then unmolded from the mold. The size of the sensor element for the current design is 10 mm in diameter and 2 mm in height. After the conductive ink was painted on its surface, the element became conductive rubber. The conductive ink is the same as the one used for making conductive fabric. The conductive rubber is similar to the conductive fabric. As a force is applied, the resistance of the rubbers will change based on the (\ref{eqn: Ohm law}). However, it is discovered that the element only reacts to certain force change~(e.g., 10~20 gw) by experiment results. There is no obvious relationship between the force and the sensing signal. The main difference is that the conductive ink only covers the surface of the rubber. That is, the cross-sectional area of the coating is quite small. If there is a large force, the rubber expands, and the cross-sectional area becomes very small and resistance surges. While this phenomenon happens, the conductive rubber is closed to an open loop, and the further change cannot be measured. That is the reason why it cannot sensitively react to the large normal force. 

However, the conductive rubber is suitable to be applied as the position-prediction layer. When a force is applied, the voltage signal increases slightly. As the force is removed, the signal goes back to zero. By the change of increasing and decreasing of the signal, the element can help to detect the contact position. In this paper, we deployed four elements in the position-detection layer. We will use more elements in the future version to increase the resolution. In addition, there are two nodes on each element, which will be connected to the amplifier circuit to get sensing information. The nodes are 3 mm by 2 mm, and its length is 19.5 mm.   

\begin{figure}[http]
    \centering
    \includegraphics[width=230pt]{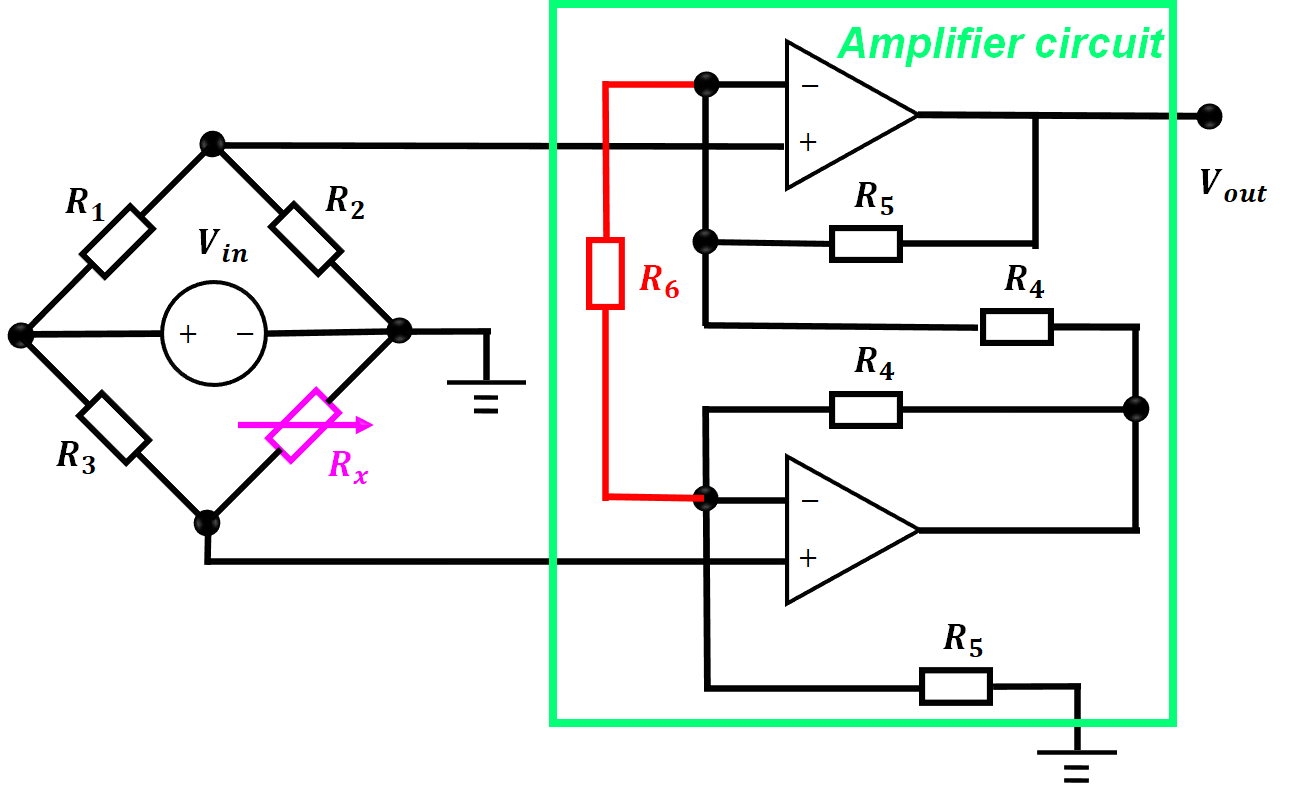}
    \caption{The Wheatstone bridge for the first layer's sensor can get the sensing signal and the amplifier circuit can amplify the signal to to further process.}
    \label{fig: wheatstone bridge for 1st layer}
\end{figure}

\subsection{Fabrication of Soft Sensor}
The terminal goal of the designed sensor is to fit in the robot grippers. In this paper, we primarily focus on its performance. The size of this sensor was determined to be 50 mm by 50 mm, and the height is 8 mm. Concerning the application purposes, the size of this design can be adjusted to meet different requirements. 
The manufacturing process is shown in Fig.~\ref{fig: process}. In the beginning, liquid rubber was poured into the mold to form the sensor's bottom layer. After the rubber cured, four elements of the position-detection layer were placed on top of the bottom layer. Then, another component of the mold was stacked on top of the bottom mold, and more liquid rubber was poured into it. Thirdly, a conductive fabric was placed on top of the sensor, and more rubber was poured. After the rubber was cured, the molds were removed, and the sensor is shown in Fig.~\ref{fig: Apparence}. For the parameters of the material we used, the resistance of the conductive fabric is approximately 100 K$\Omega$, and the elements in the second layer range from 1~M$\Omega$ to 2~M$\Omega$.

\subsection{Sensing Circuit Design}

The sensors of both layers are resistive. The commonly used sensing circuit for resistive sensors is the Wheatstone bridge~\cite{c25}. As shown in Fig.~\ref{fig: wheatstone bridge for 1st layer}, $R_x$ denotes the resistance of the sensor, and we need to choose the magnitude of $R_1$, $R_2$, and $R_3$ in order to balance the bridge, which means the current flowing out of the circuit is zero. The resistance will be selected by the rule 
\begin{align}
\frac{R1}{R2} = \frac{R3}{Rx}
\label{eqn: Wheatstone brdige}
\end{align}
Choosing all resistors have the same resistance as the sensor is one way to satisfy the above equation. To analysis the Wheatstone bridge, we firstly calculate the Thevenin resistance~\cite{c25} of the bridge,

\begin{align}
R_t = \frac {R_xR_x}{R_x+R_x} + \frac {R_x(R_x+ \Delta R_x)}{R_x+(R_x + \Delta R_x)}
\label{eqn: Wheatstone brdige ana}
\end{align}
where $R_t$ is the Thevenin resistance. The (\ref{eqn: Wheatstone brdige ana}) can be simplified as:
\begin{figure}[http]
    \centering
    \includegraphics[width=215pt]{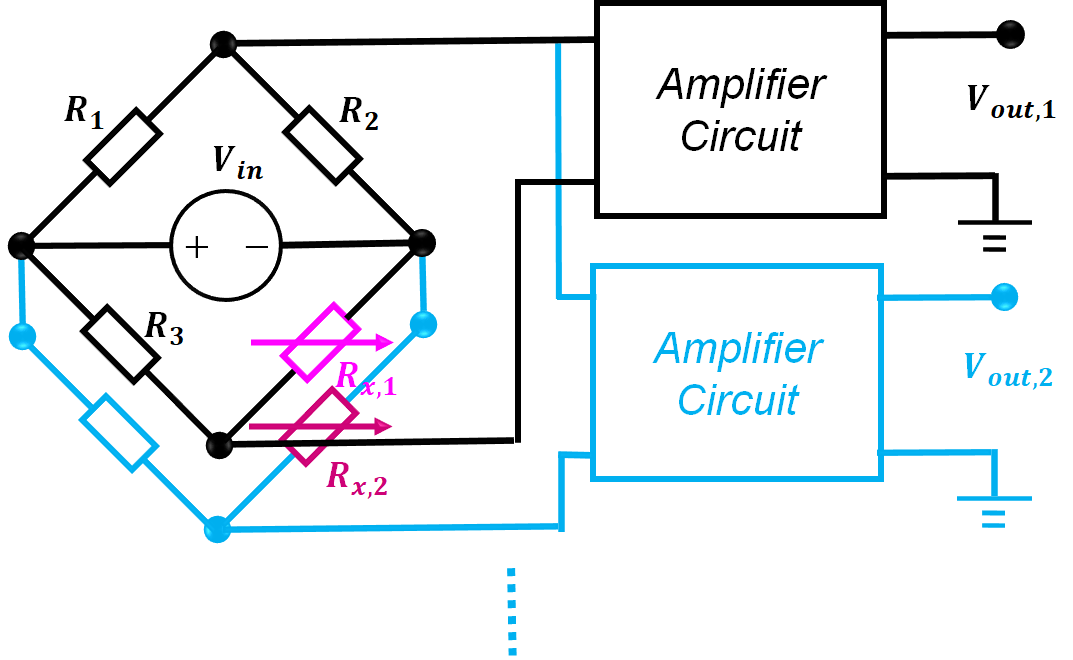}
    \caption{The Wheatstone bridge with three extended bridges were designed to get voltage signal from the four elements in the second layer.}
    \label{fig: Wheatstone bridge with extended}
\end{figure}

\begin{align}
R_t = \frac {R_x}{2} + \frac {R_x+\Delta R_x}{2R_x+\Delta R_x}{R_x}
\label{eqn: Wheatstone brdige simplified}
\end{align}
Owing to the elasticity of conductive fabric, the variation of $\Delta R_x$ can be up to 35 $\%$. The variation of $R_x$ versus $\Delta R_x$ is almost linear, and the slope change of (\ref{eqn: Wheatstone brdige simplified}) is 0.25$\sim$0.19:
This advantage can be observed via the model of the sensor in Sec III, and the experimental results in Sec IV.

Nevertheless, 
the contact force might be small and hard to discover. 
We used an INA 126A operational amplifier, which contains two operational amplifiers, to enlarge the signal. The component is precision instrumentation amplifiers and has a low offset voltage, which is able to reduce the output error. The operational amplifier also has a good ability to reject common-mode noise, which may cause by parasitic capacitance effect inside the circuit. Overall, we can obtain an accurate output without harsh noises.

\begin{figure*}[http]
    \centering
    \includegraphics[width=1\textwidth]{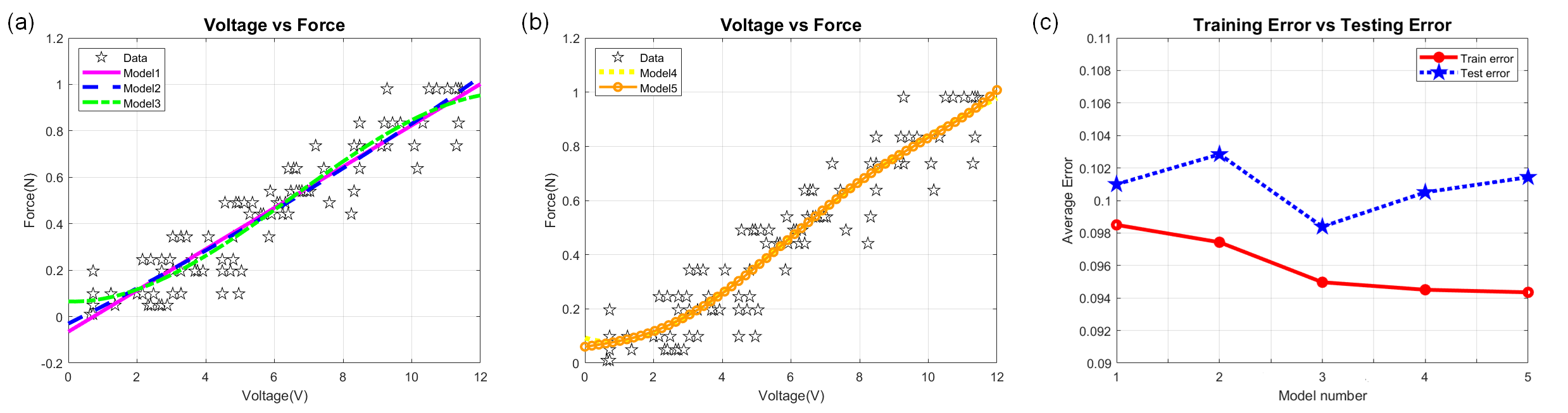}
    \caption{The modeling results of Model 1 to Model 3 can be observed in (a) while the results of Model 4 and Model 5 can be seen in (b). The collected data and averaged errors of training results can be seen in (d).}
    \label{fig: The modeling results}
\end{figure*}

We have conducted simulations on the circuit using the simulator SIMPLIS. The results indicate the sensing circuit's output errors with INA 126 amplifier are about 1$\%$. Initially, we used operational amplifier LM 741. The output errors could be up to 6$\%$, which may negatively influence the estimation of the contact force. In addition, the amplifying gain can be adjusted by tuning a single external resistor $R_6$, as shown in Fig~\ref{fig: wheatstone bridge for 1st layer}.

The circuit for the position-detection layer is similar to that of the force-measurement layer. Each element in the second layer is connected with a separate sensing circuit, which is shown in Fig.~\ref{fig: Wheatstone bridge with extended}.

\section{Sensor Modeling}

When a force is applied, the sensing signal can be measured by the sensing circuit. Then the signal is converted to the contact force via a pre-trained model.
A regression method is applied to model the transformation from the sensing signal to the contact force. For the second layer, we only need to find the threshold of each element. The threshold means the smallest force that will trigger the resistance change of the element. The thresholds were discovered by experiments.

\subsection{Collecting Data}

Collecting the data from the sensor is a necessary process to determine the model.
We used a calibration weight set to serve as the contact force since the set comes in different weights and are supposed to be more accurate. The disadvantage is that we could not get continuous data but only the discrete one. At first, we used the F/T sensor as the ground truth. The F/T sensor was placed at the bottom, and the soft sensor was placed on its top. While a force was applying, we were unable to get the correct contact force because of the sensor's weight.

Twelve different weights were used including 5, 10, 20, 25, 35, 45, 50, 55, 65, 75, 85, 100 gw. The 5, 10, 20, 50, and 100 gw are the original weights that the set has. We used two or three weights together to generate the other weights like 25 gw, 35 gw, etc. Each weight was measured eight times. Among them, the 20, 100 gw were measured nine times, and 50 gw was ten times, so we have 100 data. The data has been divided into training and testing sets. The 80 \% data is the training set, while the other 20 \% is the testing set. We utilized the k-Fold Cross Validation method to get the testing set, and k is chosen as five. The shuffled data will be split into five folds. The first fold acted as testing data, and the remaining folds served as training data.  By using k-Fold Cross-Validation, the overfitting problem of the following regression could be avoided~\cite{c28}.

\subsection{Modeling}

The common method to build a model for sensors is regression~\cite{c26,c27} since it is simple to implement. Also, the output of the sensing signal has linear behavior, as discussed in Sec II, so the regression method would be able to fit the data and build an accurate model. Before we apply this method, we need to select appropriate models. The five different models were selected to train the data and find the best model. The five models are the first-order polynomial to the fifth-order polynomial equations. The equation is expressed as
$f = a_0v + a_1 v^2 +\dots+ a_n v^{n}$. 
The $n$ stands for the order of the polynomial, $f$ is the estimated force, and $v$ is the sensing signal~(voltage).  We marked the first-order polynomial equation as Model 1, the second-order polynomial equation as Model 2, and etc.

Assume we have m training data here, and m is 80 here. If the m data is plugged into the equations, there will be m equations. Thus, they are written as matrix form and become $\bm{y}=\bm{A}\bm{x}$.

\begin{align}
\begin{bmatrix}
f_1\\
f_2\\
\vdots\\
f_m
\end{bmatrix}
=
\begin{bmatrix}
1 & v_1^{1} & \dots & v_2^{n}\\
1 & v_2^{1} & \dots & v_2^{n}\\
\vdots & \vdots & \vdots & \vdots\\
1 & v_m^{1} & \dots & v_m^{n}\\
\end{bmatrix}
\begin{bmatrix}
a_0\\
a_1\\
\vdots\\
a_n
\end{bmatrix}
\end{align}
where $\bm{y}$ $\in$ R$^{m\times1}$ is the vector that contains all force information, and $\bm{x}$ $\in$ R$^{n\times1}$ is the vector that includes all constants of the equations. The $\bm{A}$ $\in$ R$^{m\times n}$  matrix contains voltage signals from zero order to n$^{th}$ order. The objective is to minimize the 2-norm of the residual error $\bm{A}\bm{x}-\bm{y}$, i.e.,
$\min_{\substack{x}}
{||\bm{Ax}-\bm{y}||_2}$.
The equation can be solved by times $\bm{A}^{\bm{T}}$ to both side. Then, $(\bm{AA}^{\bm{T}})^{-1}$ is multiplied to both sides. The solution is 
$\bm{x}= (\bm{A}\bm{A}^{\bm{T}})^{-1}\bm{A}^{\bm{T}}y$.

We repeated the training process 20 times for each model. After eight times of training, the averaged training errors and testing errors were nearly the same. The model can be seen in Fig~\ref{fig: The modeling results}(a) and (b). The first three models are quite linear, while the last two are slightly nonlinear. It is hard to tell which model is the best by direct observation, so the averaged error was considered and could be found in Fig.~\ref{fig: The modeling results}(c). The averaged errors here are the average of repeated training results. The index used here was the root-mean-square error. The average error of the training set decreases as the model's order increases. Among the five models, the Model 3 has the smallest testing error. The testing error of Model 3 is 0.0984 N. The detailed information can be observed in Table II.  Hence, the Model 3 is the best here. In real experiments, we compared the results of Model 1 and Model 3, and no significant differences were observed between those two models. Thus, we utilize Model 1, the linear model, to estimate the contact force for this design.

\subsection{Filter Design}

\begin{table*}[h]
\caption{The averaged training and testing errors of each model}
\label{table_example}
\begin{center}\normalsize%
\begin{tabular}{|l|c|c|}
\hline
 & Training error[N] & Testing error[N]\\
\hline
Model 1 (f=-0.0650+0.0889v)  & 0.0985 & 0.1010\\
\hline
Model 2 (f=-0.0301+0.0737v+0.0012v$^{2}$) & 0.0974 & 0.1028\\
\hline
Model 3 (f=0.0653-0.0047v+0.0169v$^{2}$-0.000863v$^{3}$) & 0.0950 & 0.0984\\
\hline
Model 4 (f=0.0924-0.0405v+0.0295v$^{2}$-0.0025v$^{3}$+0.0000675v$^{4}$) & 0.0945 & 0.1005\\
\hline
Model 5 (f=0.0603+0.0189v-0.0015v$^{2}$+0.0041v$^{3}$-0.000539v$^{4}$+0.00002v$^{5}$) & 0.0944 & 0.1014\\
\hline
\end{tabular}
\end{center}%
\end{table*}

Based on our observation, the output signal from the circuit does not have severe noises which influence the estimation of force. The reason is that the operational amplifier~(INA126) can reject most noises in this design. Nevertheless, a filter was still designed for fear that there will be unexpected noises and negatively influence the estimation accuracy. The moving average filter is used here. Before the data is input into the filter, it will be converted to estimated force by using the model we built. Then, the signal will be input into the filter. The algorithm of the filter is shown below:
\begin{align}
{\overline{f} = 1/n\sum_{i=1}^{m} f(n+1-i)}
\end{align}
where $\overline{f}$ is the filtered signal, n is the number of sampled data, and $f(n+1-i)$ is the sampled data. The $m$ sampled data was considered, and $m$ is equal to 4 here. Also, each data has equal weight.

\section{Experimental Results}

\subsection{Experimental Setup}

The sensing information from the circuits is processed by using Arduino UNO R3. Arduino UNO R3 is a microcontroller based on the Microchip Atmega 325. The Arduino board is equipped with 14 digital I/O pins and six analog I/O pins. The on-chip ADC is applied to sample information from those pins, and its resolution and sampling frequency are 8 bit and 9.6 Hz, respectively. Among the analog I/O pins, one will be used to read data from the force-measurement layer. The other four will be utilized to read data from four elements from the position-detection layer. Therefore, the information will be processed concurrently, and the sensor can detect contact force and contact location simultaneously.

\begin{figure}[http]
    \centering
    \includegraphics[width=230pt]{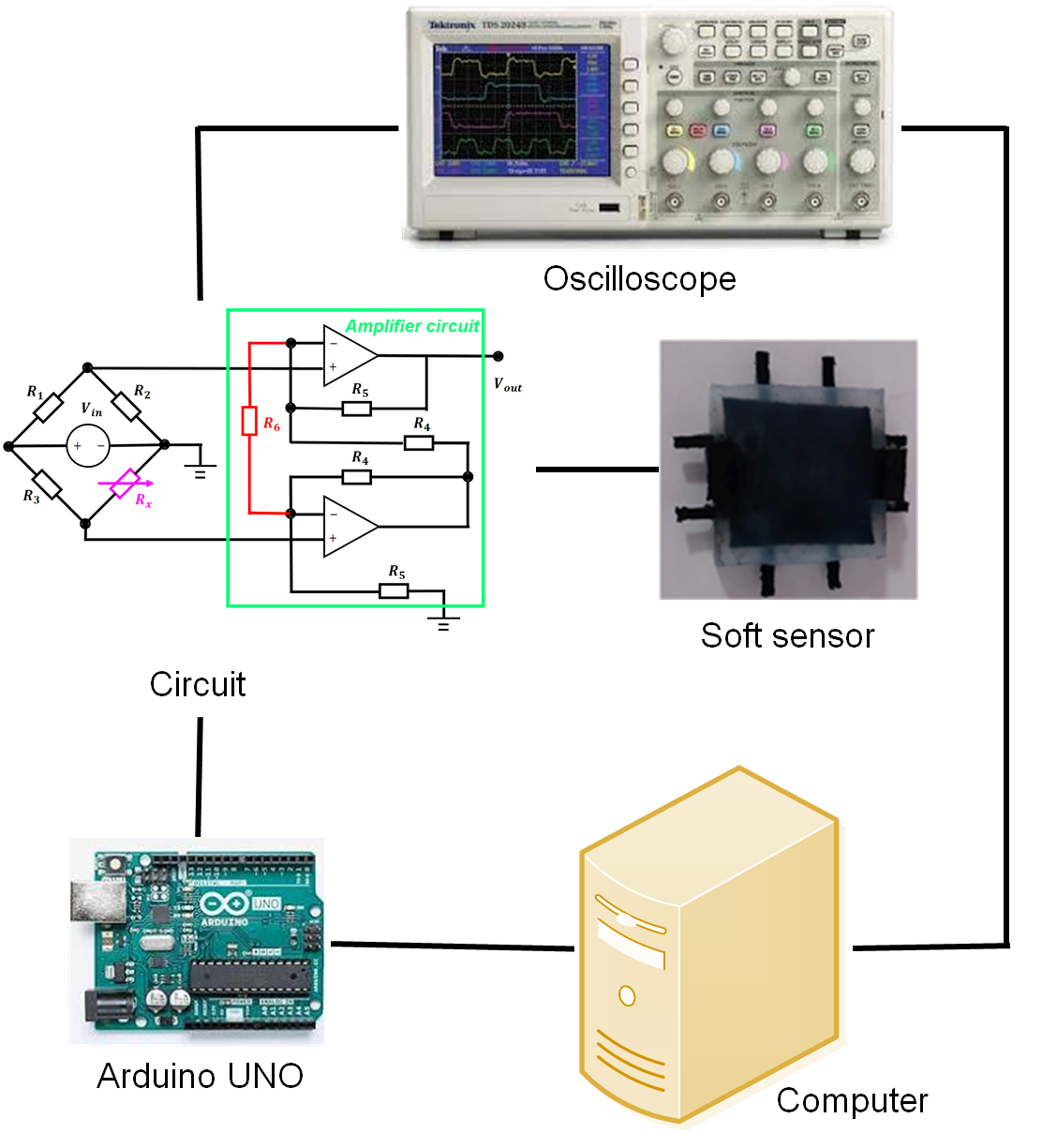}
    \caption{The experimental setup.}
    \label{fig: setup}
\end{figure}

While conducting the experiment, the sensor is connected to the circuit, as shown in Fig.~\ref{fig: setup}. The circuit is connected to both the oscilloscope and the Arduino board. The sensing signal would be processed in Arduino. The oscilloscope serves as the reference to check whether the output information from the circuit is correct. After the sensing signal is processed in Arduino, the contact force and contact position would be shown on the computer's monitor.

\subsection{Results}
\subsubsection{Thresholds of Position-detection Layer}

Before we tested the sensor, we did an experiment by using weights to find the thresholds of four elements in the position-prediction layer. These information was programmed into Arduino. Their threshols are about 0.1N. Two of them are about 0.15 to 0.2N.The differences are caused by conductive ink. The ink might not distribute averagely on the rubbers' surface, so their sensitivities are slightly different. Hence, as a force is applied and greater than the threshold of the element in that location, the sensor will simultaneously predict the contact force and contact position.

\subsubsection{Test of the Soft Sensor}

\begin{figure*}[http]
    \centering
    \includegraphics[width=1\textwidth]{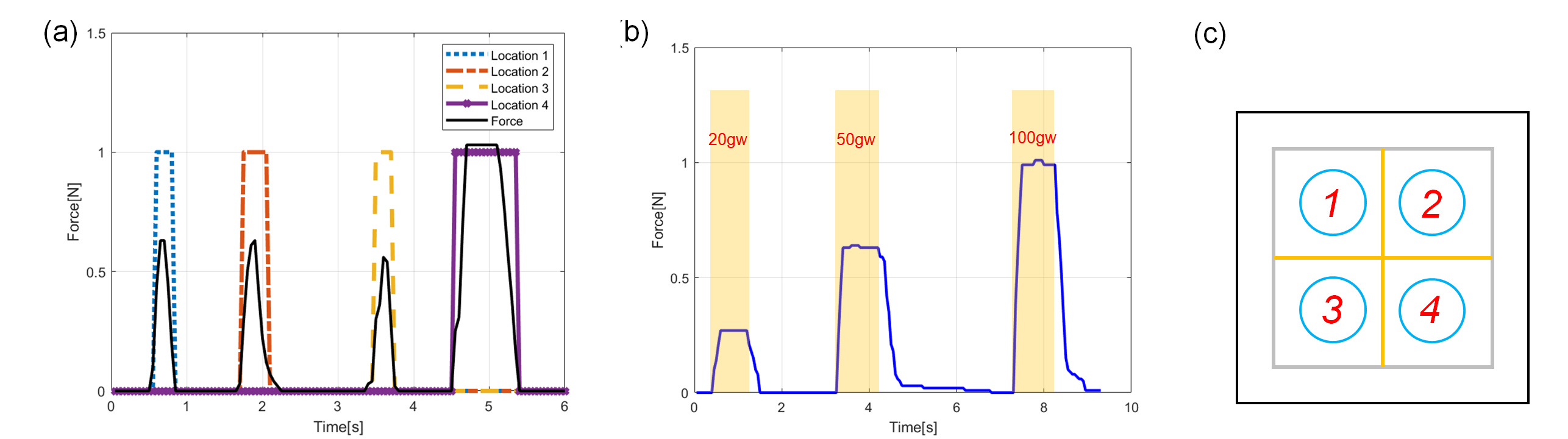}
    \caption{The experimental results can be seen in (a). The forces were applied at location 1 to 4 successively. The solid line represents the estimated contact force by the first layer, and the dotted line represents the detected contact location from the second layer. We can see from the results that both layer behaves stable in the experiments.
    Another experiment was conducted to test the sensor's accuracy. The 20, 50, and 100 gw weights were used, respectively and can be observed in (b). The soft sensor could estimate contact force and location concurrently. The locations were marked and is shown in (c).}
    \label{fig: Experimental result}
\end{figure*}

The first experiment is to know the sensor's performance. The experimental results are demonstrated in Fig.~\ref{fig: Experimental result}. The contact locations are divided into four different areas, as shown in Fig.~\ref{fig: Experimental result}(c).
We applied force on each of the areas, and the sensor was able to estimate the force and also detected the location, as shown in Fig.~\ref{fig: Experimental result}(a). In the algorithm, when the contact location was detected, it would be on mode. When the contact force reduces below the threshold of the element, it would be off mode. During the 4.5 to 5.5s in Fig.~\ref{fig: Experimental result}(a), the force was exceeding the sensing range, and the sensor hit the ceiling at 1 N. 


%

\subsubsection{Accuracy Tests of Force-Measurement Layer}
Another experiment was to test the accuracy of the force-measurement layer. The accuracy of a sensor is an important issue. Designing an accurate soft sensor for robot grippers is one of our objectives. Hence, three weights of calibration set, 20, 50, and 100 gw(0.196, 0.49, and 0.98 N), were used to test its accuracy. Those three weights were placed on the sensor successively during the experiment. The corresponding measured forces of the three weights were 0.27 N, 0.63 N, and 1.01 N. The data was sampled during the period each weight was applied on the sensor. The root-mean-square error is 0.0923 N.

\subsubsection{Sensing Range and Resolution of the Sensor}

During the experiment, the sensing range and resolution of this design were discovered. The current sensor uses the amplifier gain of 41.36, the sensing range is 1~N, and the resolution is 0.05~N. This sensor's maximum sensing range is 1.5~N, as shown in Table II with amplifier gain 22. The resolution is around 0.1~N. If we increase the amplifier gain, the resolution will become better, and the sensing range decreases. If we decrease the amplifier gain, the sensing range will increase, and the resolution becomes worse. For future applications, we can adjust the amplifier gain depends on different requirements. If the sensing range is more important than the resolution, the amplifier gain can be decreased to increases the range. If high resolution is needed, the amplifier gain will be increased to get better resolution.

\begin{table}[!h]
\caption{The sensing range and resolution of the soft sensor}
\label{table_example}
\begin{center}\normalsize%
\begin{tabular}{|c|c|c|}
\hline
Amplifier gain & Sensing ange[N] & Resolution[N]\\
\hline
22 & 1.5 & 0.1\\
\hline
41.36 & 1 & 0.05\\
\hline
\end{tabular}
\end{center}%
\end{table}

\section{CONCLUSIONS}

This paper presents a new soft sensor which is inspired by the receptors inside the fingers of humans. The sensor contains two layers of sensing units, which include a force-measurement layer and a position-detection layer.
The force-measurement layer mimics the Meissener's and Pacinian corpuscles of humans' skin, which can estimate the contact force. 
The position-detection layer imitates the Merkel's corpuscle, 
and it is able to detect the contact location. The Wheatstone bridge and amplifier circuits are utilized to get sensing information from the sensor. A model between the contact force and signal from the force-measurement layer was pre-trained by a regression method. In addition, a filter method was applied to estimate the force under noisy measurements robustly. Experiments are provided to show the performance.
For the position-detection layer, the threshold of each element was determined by experiments. The model and threshold were coded into Arduino UNO board for real-time processing. Experiment results show that the sensor is capable of measuring the contact force and contact position concurrently, and the root-mean-square error of the force-measurement layer is 0.0923~N. 

In the future, more conductive rubbers will be layout in the position-detection layer to increase its resolution and enable it to identify objects' features. The enhanced version will be deployed on a robot gripper to do dexterous manipulation.

\bibliographystyle{ieeetr}
\bibliography{IEEEabrv}

\end{document}